\documentclass{article}

\usepackage[preprint]{corl_2026}
\usepackage[T1]{fontenc}
\usepackage{amsmath,amssymb}
\usepackage{graphicx}
\usepackage{multirow}
\makeatletter
\newcommand{\captionof}[1]{\def\@captype{#1}\caption}
\makeatother

\title{PiL-World: A Chunk-Wise World Model for VLA Policy-in-the-Loop Evaluation}

\author{
  \normalfont
  Chong Ma\textsuperscript{1,2,*}\thanks{This work was completed during an internship at Midea AI Research Centers.} \quad
  Taiyi Su\textsuperscript{2,*} \quad
  Jian Zhu\textsuperscript{2,\textdaggerdbl} \quad
  Jianjun Zhang\textsuperscript{1,2} \quad
  Zitai Huang\textsuperscript{1,2} \\
  Yi Xu\textsuperscript{2} \quad
  Hanli Wang\textsuperscript{1,\textdagger} \\
  \normalfont\textsuperscript{1}Tongji University \quad
  \normalfont\textsuperscript{2}AIRC, Midea Group \\
  \normalfont\textsuperscript{*}Equal Contribution \quad
  \normalfont\textsuperscript{\textdagger}Corresponding Author \quad
  \normalfont\textsuperscript{\textdaggerdbl}Project Leader
}

\begin{document}
\maketitle

{\centering
  \vspace{-2.0em}
  Project Page: \href{https://pil-world.github.io}{pil-world.github.io}\par
  \vspace{0.6em}
}

\begin{abstract}
Vision-language-action (VLA) policies operate in a closed loop in real-world robot tasks: a robot observes the scene, executes an action chunk, and conditions its next decision on the resulting observation. However, most existing world models for robot action evaluation are limited to open-loop prediction along pre-collected action trajectories. This prevents them from supporting closed-loop VLA evaluation, where each action chunk must be conditioned on the observation generated by the previous execution. To address this gap, we propose PiL-World, a chunk-wise world model designed for \textbf{P}olicy-\textbf{i}n-the-\textbf{L}oop VLA evaluation. Given the current observation and the action chunk rolled out by a VLA policy, PiL-World generates multi-view future observations that are consistent with the VLA rollout and match the image inputs required by the policy. By alternating between VLA inference and world-model prediction, PiL-World enables closed-loop evaluation without real robot execution at every step. To improve rollout consistency, PiL-World conditions video generation on action-derived visual control from head-view robot motion and latent history memory that encodes task execution context, while jointly predicting synchronized multi-view observations. Beyond successful teleoperated demonstrations, it also learns from failed trajectories, helping the imagined rollouts better capture diverse real policy execution outcomes. We evaluate PiL-World on three real dual-arm manipulation tasks. PiL-World generates imagined rollouts that are highly consistent with real robot executions. More importantly, compared with the baseline, it reduces the error between VLA success rates measured in real-world rollouts and those estimated through closed-loop world-model evaluation from \textbf{63.2\%} to \textbf{12.0\%}.
\end{abstract}


\section{Introduction}

Vision-language-action (VLA) models have become a promising approach for general-purpose robotic manipulation~\citep{brohan2022rt1,brohan2023rt2,openx2023rtx,kim2024openvla,black2025pi0}. In real robot deployment, VLA policies are evaluated in a closed loop rather than as a single open-loop action sequence. At each step, the robot observes the scene, executes a predicted action chunk, and conditions its next prediction on the resulting observation. This observe-act feedback loop shapes the distribution of states and actions encountered during deployment, making closed-loop testing essential for reliable VLA evaluation. However, real-robot evaluation is costly and difficult to scale because it is limited by hardware safety, scene resets, and experimental throughput.

\begin{figure}[t]
  \centering
  \includegraphics[width=\linewidth]{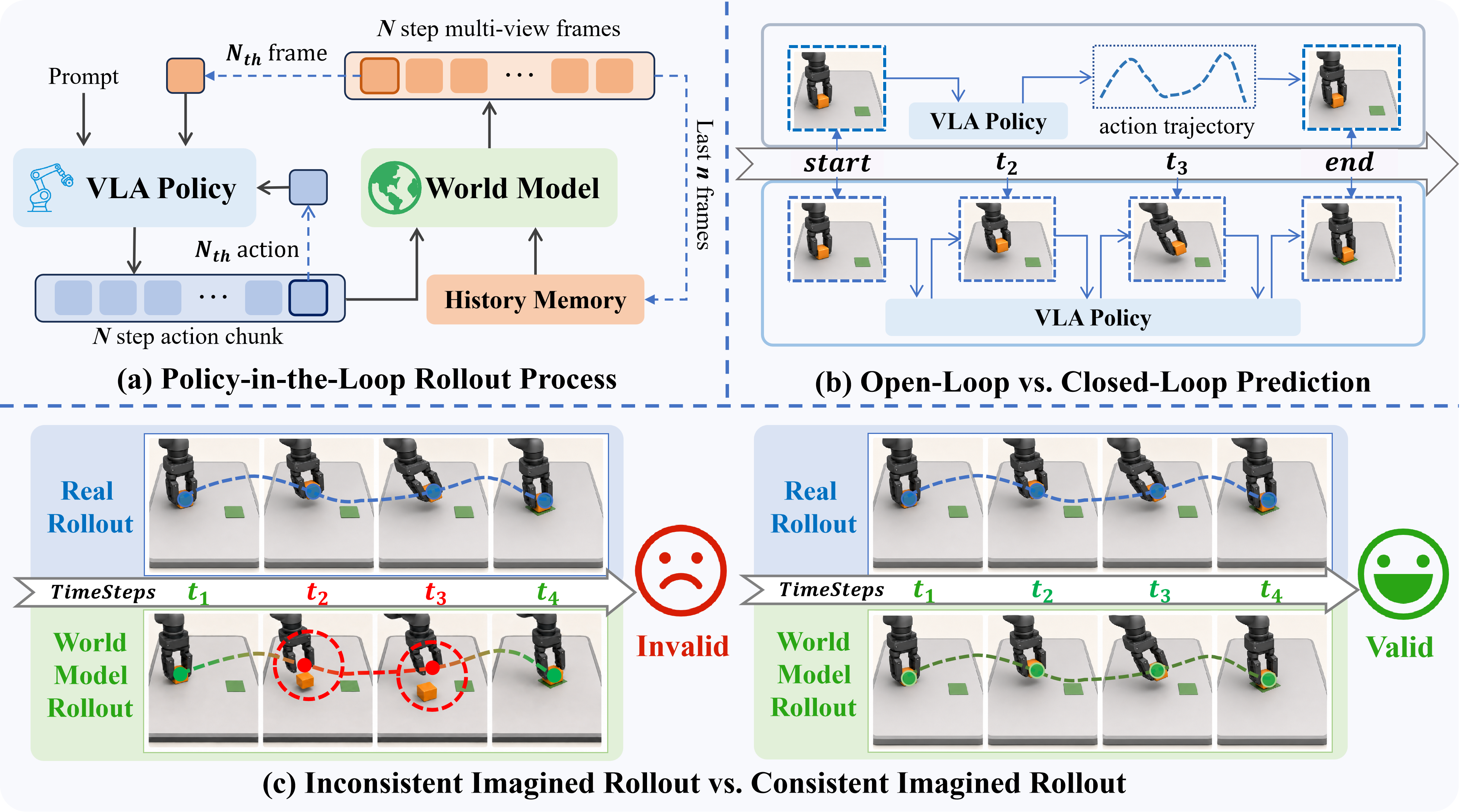}
  \caption{Motivation for policy-in-the-loop world-model evaluation.
\textbf{(a) Policy-in-the-Loop Rollout Process:} The world model predicts future observations from VLA action chunks and feeds them back for policy queries.
\textbf{(b) Open-Loop vs. Closed-Loop Prediction:} Closed-loop evaluation requires intermediate generated observations to update the policy input, whereas open-loop prediction follows a fixed action sequence and evaluates the predicted outcome without re-querying the policy.
\textbf{(c) Inconsistent Imagined Rollout vs. Consistent Imagined Rollout:} A valid imagined rollout should remain consistent with the real execution process.}
  \label{fig:motivation}
\end{figure}

World models offer a way to approximate real-robot closed-loop interaction through imagined rollouts. However, most
existing world models for robot action evaluation focus on open-loop prediction along pre-collected action trajectories,
rather than evaluating policies that repeatedly observe, act, and re-plan. As illustrated in Fig.~\ref{fig:motivation}(a), policy-in-the-loop evaluation requires the world model to map VLA-predicted action chunks to future observations and feed them back for subsequent policy queries. This differs from open-loop prediction in Fig.~\ref{fig:motivation}(b), where the action trajectory is pre-collected and fixed, so the world model predicts along the same action sequence without re-querying the policy from intermediate generated observations. Beyond this interface mismatch, task-level success alone is insufficient to validate imagined rollouts: as shown in Fig.~\ref{fig:motivation}(c), an imagined rollout may be unreliable if the generated process deviates from the real execution or exhibits multi-view inconsistency. Therefore, a gap remains between existing world-model evaluation protocols and the closed-loop, process-consistent interface required for policy-in-the-loop VLA evaluation.

We propose PiL-World, a chunk-wise world model for closed-loop VLA evaluation. PiL-World generates future observations from VLA-predicted action chunks and feeds the terminal generated observation back for subsequent policy queries. PiL-World uses (1) frame-aligned visual control signals from head-view robot motion, (2) latent multi-view history conditioning, and (3) joint multi-view prediction to generate temporally consistent future observations~\citep{rigter2024avid,su2025mtvworld}. To obtain both general dynamics priors and task-specific execution knowledge, PiL-World is trained in two stages: it is first pretrained on RealSource World~\citep{realsourceworld2025} to learn general robot-environment dynamics, and then fine-tuned on target-task executions that include both successful demonstrations and failed trajectories. During inference, PiL-World predicts one chunk of future observations, feeds the terminal observation back to the VLA policy, updates the history memory and policy-side proprioceptive state, and repeats this process to emulate real-robot closed-loop evaluation.

We evaluate PiL-World on three real dual-arm manipulation tasks: sorting cubes, stacking bowls, and stacking blocks. Using the same initial observations, task instructions, and VLA policies, we compare PiL-World closed-loop rollouts with real-robot executions to assess how well imagined rollouts reflect real policy behavior. For rollout-level evaluation, we measure task-level success agreement and introduce a new hallucination-free ratio (HFR), which quantifies how long dense imagined rollouts remain credible before obvious hallucination appears. We further evaluate single-step multi-view perceptual similarity to assess local action-conditioned generation quality.

Our contributions are threefold:
\begin{itemize}
    \setlength\itemsep{0pt}
    \setlength\parsep{0pt}
    \setlength\topsep{2pt}
    \item We introduce PiL-World, a chunk-wise world model for policy-in-the-loop VLA evaluation that matches the closed-loop interface of real-robot testing.
    \item PiL-World integrates action-derived visual control, latent multi-view history conditioning, joint multi-view prediction, and success/failure trajectory training, enabling highly consistent imagined rollouts in closed-loop evaluation.
    \item We validate PiL-World on three real dual-arm manipulation tasks and introduce hallucination-free ratio (HFR) for dense rollout reliability, showing closer agreement with real-world VLA success rates.
\end{itemize}

\section{Related Work}

\subsection{World Models for Policy Evaluation}

Vision-language-action (VLA) policies such as RT-1~\citep{brohan2022rt1}, RT-2~\citep{brohan2023rt2}, OpenVLA~\citep{kim2024openvla}, and $\pi_0$~\citep{black2025pi0} are typically evaluated through real-robot closed-loop rollouts, but such evaluation is costly and difficult to scale. To reduce this cost, prior work has explored simulation and world models as alternatives to repeated real-robot trials. SimplerEnv~\citep{li2024simpler} evaluates robot policies in simulation, while WorldEval~\citep{li2025worldeval}, dWorldEval~\citep{li2026dworldeval}, and WorldGym~\citep{quevedo2025worldgym} use generated or simulated rollouts to estimate policy performance. Other works use imagined interactions for robot learning or behavior simulation, such as DayDreamer~\citep{wu2023daydreamer} and IRASim~\citep{zhu2024irasim}. More recent systems, including Ctrl-World~\citep{guo2026ctrlworld} and Hi-WM~\citep{li2026hiwm}, further couple world models with policies for rollout-based evaluation or post-training. Together, these works show that imagined rollouts can provide useful evidence before real-robot testing. PiL-World builds on this direction and focuses on the closed-loop interface used in VLA evaluation: the policy predicts an action chunk, the world model generates the corresponding stride-aligned future observations, and the generated observations are fed back for the next policy query.

\subsection{Action-Conditioned Robot Video Prediction}

Policy-in-the-loop evaluation requires a world model to turn VLA-predicted actions into future observations that can be used for later policy queries. Early visual foresight and action-conditioned prediction methods showed that future frames can support robotic decision making~\citep{oh2015action,xie2019visualforesight}. Diffusion-based video generation models, often operating in latent space, later improved the quality of conditional visual generation~\citep{ho2020ddpm,rombach2022ldm,blattmann2023svd}, making them useful backbones for robot world models. Recent robot video models study different ways to condition generation on actions: AVID~\citep{rigter2024avid} adapts video diffusion models to action-conditioned robot prediction, and MTV\mbox{-}World~\citep{su2025mtvworld} converts robot control signals into visual control videos to improve multi-view consistency. WorldVLA~\citep{worldvla2025} and VLAW~\citep{guo2026vlaw} further connect world models with VLA policies for prediction, representation learning, or iterative improvement, while ABot-PhysWorld~\citep{chen2026abotphysworld} studies physical realism and action alignment in embodied world models. PiL-World follows this action-conditioned prediction setting, but uses action-derived visual control and latent multi-view history conditioning in a policy-in-the-loop rollout setting, where generated observations are fed back to form the next policy input.

\section{PiL-World}

\subsection{Problem Definition}

We study world-model rollout for VLA policy-in-the-loop evaluation. Let $\mathcal{V}$ be the set of camera views, $\mathbf{x}_t=\{x_t^v\}_{v\in\mathcal{V}}$ the multi-view observation at time $t$, $\mathbf{s}_t$ the robot proprioceptive state, and $g$ the task instruction. Then a VLA policy $\pi$ predicts an action chunk $A_t$ of horizon $H_\pi$:
\[
A_t=\{a_{t+1},a_{t+2},\ldots,a_{t+H_\pi}\}
= \pi(\mathbf{x}_t,\mathbf{s}_t,g).
\]
In policy-in-the-loop evaluation, the world model takes the action chunk and generates future observations needed for the next policy query. Let $W_\theta$ denote PiL-World, and let $\mathcal{H}_t$ denote the latent history memory at time $t$. For a prediction horizon of $K$ frames with stride $\Delta$, where $K\Delta\le H_\pi$, we use the stride-aligned actions
\[
A_t^{\Delta,K}=\{a_{t+\Delta},a_{t+2\Delta},\ldots,a_{t+K\Delta}\},
\]
and map them into a visual control condition with $\Gamma$, where $\Gamma$ denotes the deterministic action-to-control projection introduced below. PiL-World predicts one imagined execution segment as
\[
\hat{\mathbf{x}}_{t+\Delta:t+K\Delta:\Delta}
\sim
W_\theta\!\left(
\cdot \mid
\mathbf{x}_t,\mathcal{H}_t,\Gamma(A_t^{\Delta,K}),g
\right).
\]
The generated observations form one imagined execution segment. By feeding the terminal generated observation back to the policy and updating the history memory, PiL-World composes these segments into a closed-loop imagined rollout.

\begin{figure}[t]
\centering
\includegraphics[width=\linewidth]{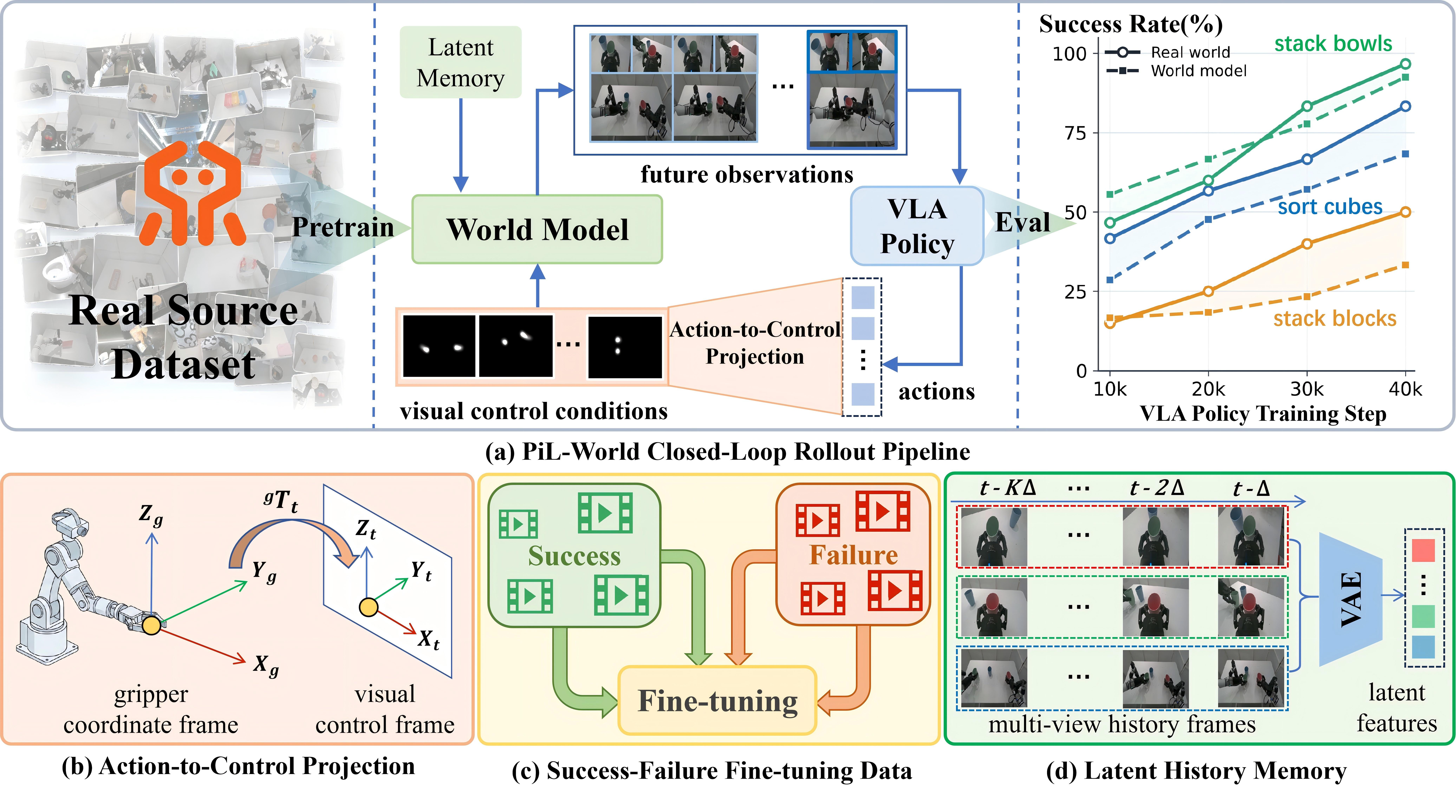}
\caption{Overview of PiL-World.
\textbf{(a) PiL-World closed-loop rollout pipeline:} PiL-World uses action chunks predicted by a VLA policy to generate a stride-aligned future observation segment. The terminal generated observation is fed back for the next policy query, enabling imagined rollouts to be compared with real-robot outcomes across policy checkpoints.
\textbf{(b) Action-to-control projection:} Policy-predicted actions are projected into visual gripper-motion control signals for head-view action conditioning.
\textbf{(c) Success/failure fine-tuning data:} PiL-World is fine-tuned on a mixture of successful demonstrations and failed teleoperated executions, exposing the world model to both goal-reaching and non-goal-reaching trajectories.
\textbf{(d) Latent history memory:} Recent multi-view frames are encoded into history latents that condition future prediction and preserve rollout context.}
\label{fig:method}
\end{figure}

\subsection{Framework}

Figure~\ref{fig:method} illustrates the full PiL-World pipeline from training to policy-in-the-loop evaluation. PiL-World first learns general robot-environment dynamics through pretraining on RealSource World~\citep{realsourceworld2025}, and is then fine-tuned on target-task trajectories containing both successful demonstrations and failed teleoperated executions. During rollout, a frozen VLA policy predicts an action chunk from the current observation and instruction. PiL-World projects this action chunk into visual control signals, conditions generation on latent history memory, and predicts a stride-aligned multi-view future segment. The terminal generated observation is fed back for the next policy query, and the resulting imagined rollout is compared with real-robot execution for evaluation.

\paragraph{PiL-World Closed-Loop Rollout Pipeline.}
At inference time, PiL-World alternates with a frozen VLA policy for rollout rounds $r=1,\ldots,R$. In each round, the policy predicts an action chunk from the current imagined observation, policy-side proprioceptive state, and task instruction. PiL-World converts the chunk into stride-aligned visual control conditions and generates the next $K$ multi-view frames. The terminal generated observation becomes the next policy observation, while recent frames, real at initialization and generated thereafter, are re-encoded as latent history memory. Since actions are absolute joint-space commands, the policy-side proprioceptive state is updated from the last stride-aligned action. Repeating this process yields a dense imagined trajectory covering up to $RK\Delta$ original time steps.

\paragraph{Action-to-Control Projection.}
PiL-World conditions future prediction on actions produced by the VLA policy. Robot actions are absolute joint-space commands, while the video model mainly operates on visual inputs. We therefore convert the actions into head-view visual control frames, using the control-video representation~\citep{su2025mtvworld}.

Specifically, we map stride-aligned robot actions into a head-view visual control sequence aligned with future frames.
As shown in Fig.~\ref{fig:method}(b), we convert absolute dual-arm action commands into head-view gripper
markers through robot kinematics and camera projection. The marker locations encode projected gripper positions, while
marker sizes encode the corresponding gripper states. We denote this deterministic projection process as
\[
C_t
=
\Gamma(A_t^{\Delta,K})
=
\{c_{t+\Delta},c_{t+2\Delta},\ldots,c_{t+K\Delta}\}.
\]
Each control frame $c_{t+k\Delta}$ is aligned with action $a_{t+k\Delta}$ and serves as a visual
proxy for gripper motion in the head view. The geometric projection procedure is described in Appendix~\ref{app:control_video}.

\paragraph{Success/Failure Fine-Tuning Data.}
PiL-World is trained to capture both general robot dynamics and task-specific execution outcomes. We first pretrain the model on RealSource World~\citep{realsourceworld2025} to learn general action-conditioned robot-environment dynamics, and then fine-tune it on target-task trajectories collected from the same real-robot setup used for evaluation. As shown in Fig.~\ref{fig:method}(c), the target-task data include both successful demonstrations and failed teleoperated executions, exposing the model to goal-reaching and non-goal-reaching trajectories.

\paragraph{Latent History Memory.}
In closed-loop rollout, prediction is performed over multiple rounds, and each new segment is generated from observations produced in previous rounds. Without recent visual context, the generated segment may drift away from the preceding rollout history. Therefore, as illustrated in Fig.~\ref{fig:method}(d), PiL-World maintains a latent history memory $\mathcal{H}_t$ by encoding recent multi-view frames as visual context for future prediction. These history frames are real observations at the beginning of rollout and are updated with generated observations in later rollout rounds. Let $\mathcal{I}_t^h$ denote the indices of recent history frames. With the VAE encoder $E_{\phi}$, for each view $v\in\mathcal{V}$ we encode the current frame, history frames, and future target frames as
\[
Z_t^{v,0}=E_\phi(x_t^v),
\qquad
Z_t^{v,h}=E_\phi(\{x_\tau^v\}_{\tau\in\mathcal{I}_t^h}),
\qquad
Z_t^{v,f}=E_\phi(\{x_\tau^v\}_{\tau=t+\Delta:t+K\Delta:\Delta}).
\]
The latent history memory is defined as $\mathcal{H}_t=\{Z_t^{v,h}\}_{v\in\mathcal{V}}$, and the future target latents are $Z_t^f=\{Z_t^{v,f}\}_{v\in\mathcal{V}}$. During training, the current-frame latents, latent history memory, visual control condition, and instruction are used as conditions, while denoising supervision is applied only to $Z_t^f$. At inference time, PiL-World uses $\mathcal{H}_t$ to condition future-latent generation, and the generated latents are decoded into synchronized multi-view observations. This design preserves temporal context across rollout rounds, and the latent denoising objective is detailed in Appendix~\ref{sec:latent_objective}.

\subsection{Training and Evaluation Alignment}

During training, to match the closed-loop rollout interface above, long robot trajectories are split into action-conditioned clips, each corresponding to one world-model prediction step. Each clip contains recent multi-view history frames, the current observation, stride-aligned action conditions, and future observations sampled at the prediction stride. During inference, multiple prediction steps are composed into a complete imagined rollout. For evaluation, imagined and real branches are initialized from the same subtask observation and task instruction, using the same VLA policy. This paired protocol tests whether imagined rollouts follow real-environment behavior under matched policy inputs, beyond producing visually plausible videos. The detailed rollout construction, annotation protocol, and metrics are provided in Sec.~\ref{sec:evaluation_metrics} and Appendix~\ref{app:metrics}.

\section{Experiments}

\subsection{Experimental Setup}

\noindent\textbf{Tasks and policy evaluation setup.}
We evaluate PiL-World on three real dual-arm manipulation tasks: sorting cubes, stacking bowls, and stacking blocks, with task definitions and language instructions in Appendix~\ref{app:task_descriptions}. We use $\pi_0$~\citep{black2025pi0} as the VLA policy. The policy is fine-tuned only on successful demonstrations, while PiL-World is fine-tuned on both successful demonstrations and failed teleoperated trajectories. The training/test splits contain $100/20$, $100/18$, and $200/20$ full episodes for sorting cubes, stacking bowls, and stacking blocks, respectively. Evaluation is performed at the subtask level, yielding $60$, $54$, and $40$ test subtasks, respectively. For each test subtask, real and imagined branches start from the same observation and instruction; the real branch executes a frozen VLA checkpoint on the robot, while the imagined branch replaces robot dynamics with PiL-World and feeds terminal generated observations back to the policy.

\noindent\textbf{World-model training and baseline setup.}
We initialize $W_\theta$ from Wan2.1-14B~\citep{wan2025wan} and further pretrain it on RealSource World~\citep{realsourceworld2025}, a large-scale real-world dual-arm manipulation dataset, to adapt the model to robot-environment dynamics. RealSource World contains over 14 million frames from 11,428 episodes spanning 35 manipulation tasks. We then fine-tune the model on target-task trajectory clips using LoRA~\citep{hu2022lora}. Each training clip corresponds to one world-model prediction step. Each step predicts $K=15$ future frames with stride $\Delta=3$, and clips are extracted using a sliding-window stride of $9$ stride-sampled frames. 
Ctrl-World~\citep{guo2026ctrlworld}, a state-of-the-art policy-compatible world model for imagined policy rollouts, serves as our primary baseline. We compare with Ctrl-World under a horizon-matched protocol using the same target-task split, subtask starts, total predicted horizon, and evaluation granularity. Additional implementation details are provided in Appendix~\ref{app:baseline}.

\subsection{Evaluation Metrics}
\label{sec:evaluation_metrics}
We evaluate PiL-World using three metrics. First, real--imagined success agreement is measured by the absolute success-rate gap $|\Delta\mathrm{SR}|=|\mathrm{SR}_{\mathrm{imag}}-\mathrm{SR}_{\mathrm{real}}|$, where $\mathrm{SR}_{\mathrm{imag}}$ and $\mathrm{SR}_{\mathrm{real}}$ denote imagined and real-robot success rates, respectively; Pearson correlation is additionally reported across task-checkpoint settings. Second, we introduce hallucination-free ratio (HFR), a human-annotated metric that measures the normalized position of the first obvious hallucination in dense imagined rollouts; higher HFR indicates that the rollout remains credible for a larger fraction of the predicted trajectory. Third, single-step LPIPS~\citep{zhang2018lpips} measures short-horizon multi-view perceptual similarity over one world-model prediction step under ground-truth action conditioning. Full metric definitions and annotation rules are provided in Appendix~\ref{app:metrics}.

\begin{table}[!tbp]
\centering
\begin{minipage}{0.49\linewidth}
\centering
\vspace{0pt}
\captionof{table}{Real-world-aligned closed-loop rollout for the 40k-step VLA checkpoint. All values are percentages. Lower $|\Delta\mathrm{SR}|$ indicates closer agreement with real-robot success, and higher HFR indicates longer hallucination-free rollouts.}
\label{tab:rollout_results}
\vspace{0.2em}
\footnotesize
\begingroup
\renewcommand{\arraystretch}{1.04}%
\setlength{\tabcolsep}{3pt}%
\begin{tabular*}{\linewidth}{@{\extracolsep{\fill}}lccc@{}}
\hline
\multirow{2}{*}{Method} & \multirow{2}{*}{$\mathrm{SR}_{\mathrm{imag}}$} & \multirow{2}{*}{$\lvert\Delta\mathrm{SR}\rvert$ $\downarrow$} & \multirow{2}{*}{HFR $\uparrow$} \\ \\
\hline
& \multicolumn{3}{c}{\textit{Sort Cubes} ($\mathrm{SR}_{\mathrm{real}}=83.3$)} \\
\cline{2-4}
Ctrl-World & 11.5 & 71.8 & 39.5 \\
PiL-World  & 68.3 & \textbf{15.0} & \textbf{83.3} \\
\hline
& \multicolumn{3}{c}{\textit{Stack Bowls} ($\mathrm{SR}_{\mathrm{real}}=96.7$)} \\
\cline{2-4}
Ctrl-World & 24.1 & 72.6 & 47.4 \\
PiL-World  & 92.5 & \textbf{4.2} & \textbf{83.9} \\
\hline
& \multicolumn{3}{c}{\textit{Stack Blocks} ($\mathrm{SR}_{\mathrm{real}}=50.0$)} \\
\cline{2-4}
Ctrl-World & 4.9 & 45.1 & 37.7 \\
PiL-World  & 33.3 & \textbf{16.7} & \textbf{43.0} \\
\hline
\end{tabular*}
\endgroup
\end{minipage}
\hfill
\begin{minipage}{0.49\linewidth}
\centering
\vspace{0pt}
\captionof{table}{Single-step visual prediction under ground-truth action conditioning. We report average multi-view LPIPS over the 15-frame prediction horizon of a single world-model prediction step; lower values indicate better perceptual similarity.}
\label{tab:lpips_results}
\vspace{0.2em}
\footnotesize
\begingroup
\renewcommand{\arraystretch}{1.04}%
\setlength{\tabcolsep}{0pt}%
\begin{tabular*}{\linewidth}{@{\extracolsep{\fill}}cccc@{}}
\hline
\multirow{2}{*}{Method} & Overall & Head & Wrist Avg. \\
 & LPIPS $\downarrow$ & LPIPS $\downarrow$ & LPIPS $\downarrow$ \\
\hline
 & \multicolumn{3}{c}{\textit{Sort Cubes}} \\
\cline{2-4}
Ctrl-World & 0.1454 & 0.1030 & 0.1666 \\
PiL-World & \textbf{0.0965} & \textbf{0.0597} & \textbf{0.1148} \\
\hline
 & \multicolumn{3}{c}{\textit{Stack Bowls}} \\
\cline{2-4}
Ctrl-World & 0.1366 & 0.0959 & 0.1569 \\
PiL-World & \textbf{0.1100} & \textbf{0.0597} & \textbf{0.1351} \\
\hline
 & \multicolumn{3}{c}{\textit{Stack Blocks}} \\
\cline{2-4}
Ctrl-World & 0.1277 & 0.0885 & \textbf{0.1474} \\
PiL-World & \textbf{0.1208} & \textbf{0.0617} & 0.1503 \\
\hline
\end{tabular*}
\endgroup
\end{minipage}
\end{table}

\begin{figure}[!tbp]
\centering
\begin{minipage}[t]{0.48\linewidth}
\centering
\includegraphics[width=1\linewidth]{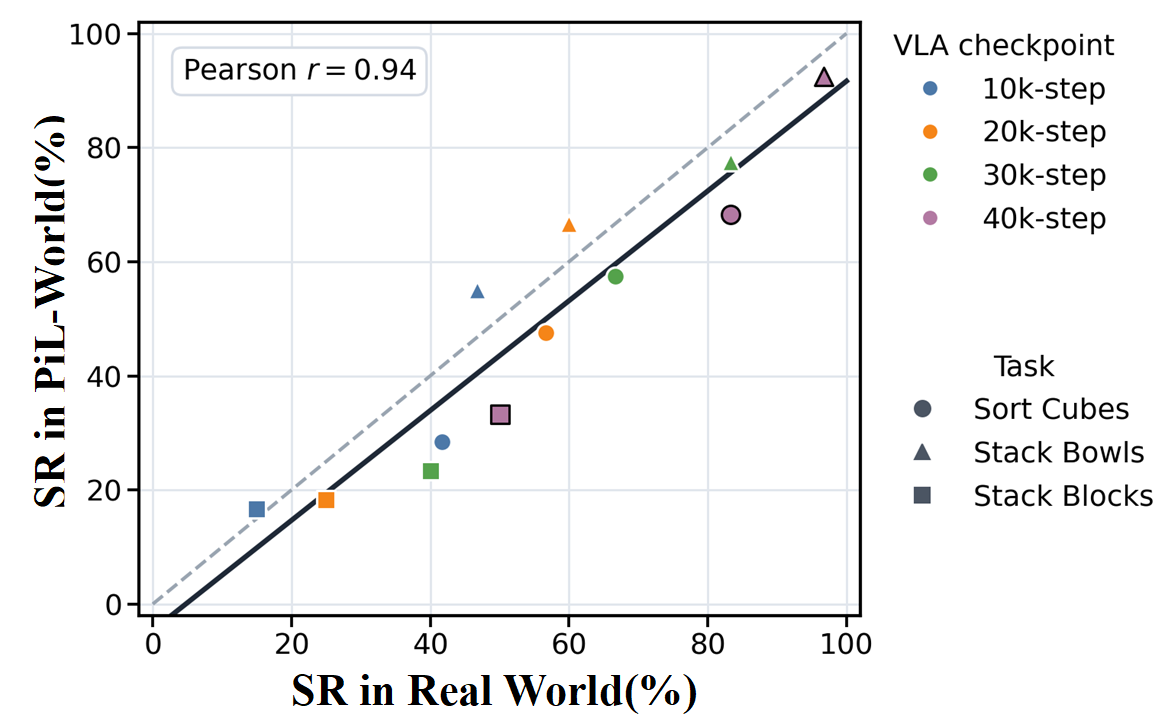}
\captionof{figure}{Real--imagined success agreement across VLA checkpoints. Each point is a task--checkpoint setting; shape denotes task and color denotes checkpoint. The trend indicates whether imagined rollouts preserve real-world policy performance across checkpoints.}
\label{fig:rollout_success_trend}
\end{minipage}
\hfill
\begin{minipage}[t]{0.48\linewidth}
\centering
\includegraphics[width=0.7\linewidth]{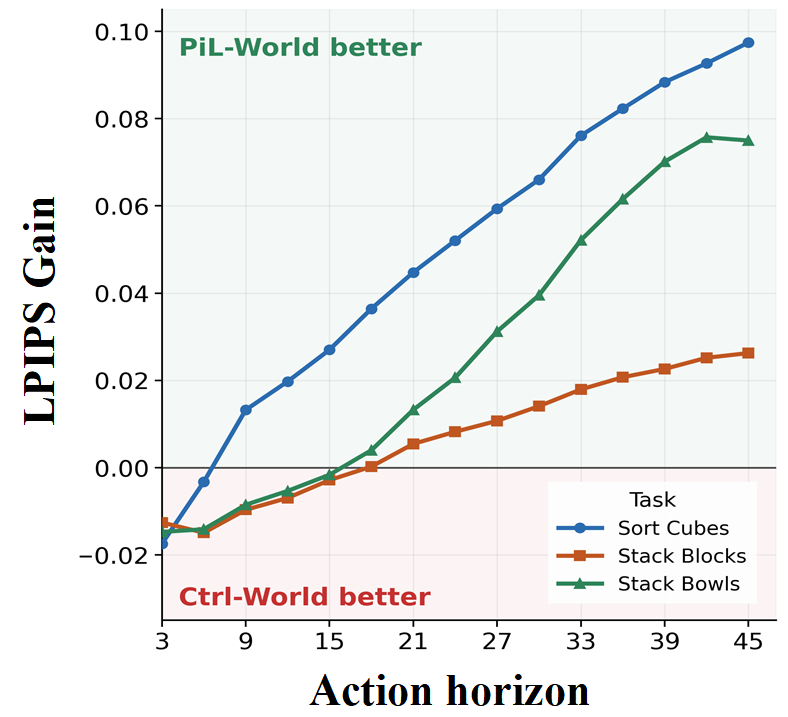}
\captionof{figure}{Action-horizon LPIPS gain under ground-truth action conditioning. We compute the gain as Ctrl-World
LPIPS minus PiL-World LPIPS at each predicted frame; positive values indicate lower LPIPS for PiL-World.}
\label{fig:lpips_horizon}
\end{minipage}
\end{figure}

\subsection{Main Results}
\textbf{Closed-loop rollout agreement.}
Table~\ref{tab:rollout_results} reports real-world-aligned closed-loop rollout results under the horizon-matched protocol. Across the three tasks, PiL-World reduces the average real--imagined success-rate gap from $63.2\%$ for Ctrl-World to $12.0\%$. This improvement is consistent across task types: PiL-World substantially reduces the gap on Sort Cubes, closely matches real success on Stack Bowls, and also narrows the gap on the more contact-sensitive Stack Blocks task. These results indicate that PiL-World does not merely generate visually plausible rollouts, but better preserves the task-level outcomes induced by the same frozen VLA policy. HFR complements this result by measuring how long the generated rollout remains credible before obvious hallucination appears. PiL-World improves average HFR from $41.5\%$ to $70.1\%$, with large gains on Sort Cubes and Stack Bowls. The smaller gain on Stack Blocks is consistent with the difficulty of stable contact-rich stacking, where small pose errors can quickly amplify over a closed-loop rollout.

Figure~\ref{fig:rollout_success_trend} evaluates real--imagined success agreement across multiple VLA checkpoints. PiL-World achieves a Pearson correlation of $0.94$ between real and imagined success rates, indicating strong real--imagined agreement across task--checkpoint settings. This suggests that PiL-World can reflect relative policy performance, making imagined rollouts more useful as an evaluation proxy rather than only qualitative visualization.

\textbf{Single-step visual prediction.}
Table~\ref{tab:lpips_results} reports the average LPIPS over the 15 frames predicted by a single world-model prediction step under ground-truth action conditioning, isolating local action-conditioned visual prediction quality. PiL-World achieves lower overall LPIPS than Ctrl-World on all three tasks, with relative reductions of $33.7\%$, $19.5\%$, and $5.4\%$ on Sort Cubes, Stack Bowls, and Stack Blocks, respectively. The largest gains appear on the head view, where the action-to-control projection directly constrains gripper motion, while the wrist-view average is slightly worse on Stack Blocks. This pattern suggests that PiL-World improves the action-aligned view used to guide future prediction, while wrist-view prediction remains challenging under close-range occlusion and contact. An ablation under the same single-step LPIPS protocol further shows that removing latent history memory substantially increases perceptual prediction error across all three tasks, highlighting the role of recent multi-view context in preserving visual consistency; see Appendix~\ref{app:history_ablation}.

Figure~\ref{fig:lpips_horizon} further shows that the LPIPS gain becomes more pronounced at later predicted frames. This is important for policy-in-the-loop evaluation because each rollout step should provide observations that remain useful until the next policy query. Positive gains at later predicted frames indicate that PiL-World maintains lower perceptual prediction error over longer action-conditioned horizons.

\noindent\textbf{Qualitative rollout consistency.}
Figure~\ref{fig:rollout_examples} compares rollout examples from the same initial observation. The red dashed boxes highlight regions where the generated rollout becomes visibly inconsistent with the reference execution. Such inconsistencies are consequential in closed-loop evaluation because they can change the visual evidence provided to the policy and affect subsequent action predictions. Compared with Ctrl-World, PiL-World better preserves rollout consistency over the predicted segment, supporting the HFR results and the need for closed-loop rollout reliability in VLA evaluation.

\begin{center}
\centering
\begin{minipage}[t]{0.49\linewidth}
\centering
\vspace{0pt}
\includegraphics[width=\linewidth]{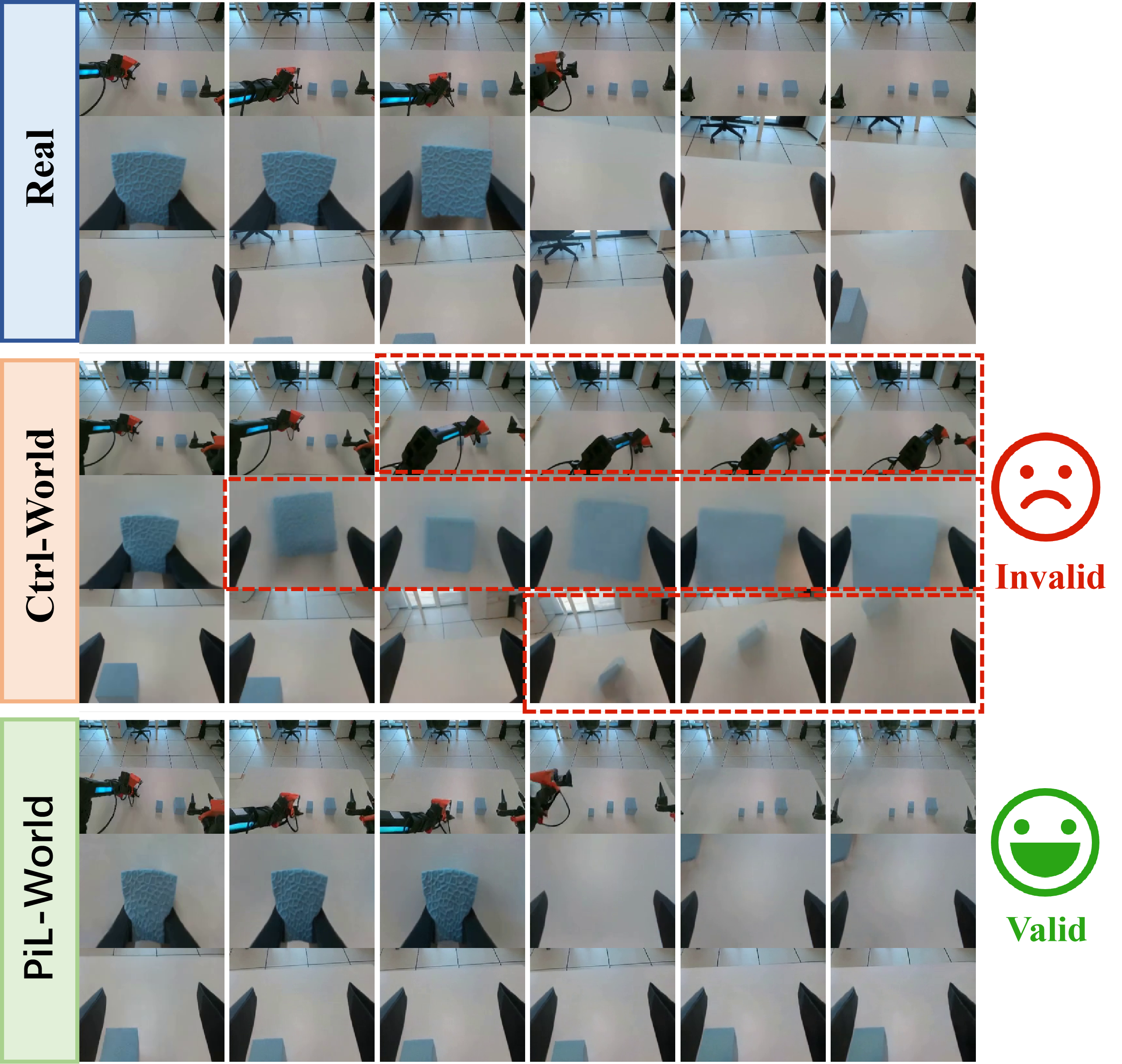}\\[-0.2em]
{\small (a) Rollout example on Sort Cubes}
\end{minipage}
\hfill
\begin{minipage}[t]{0.49\linewidth}
\centering
\vspace{0pt}
\includegraphics[width=\linewidth]{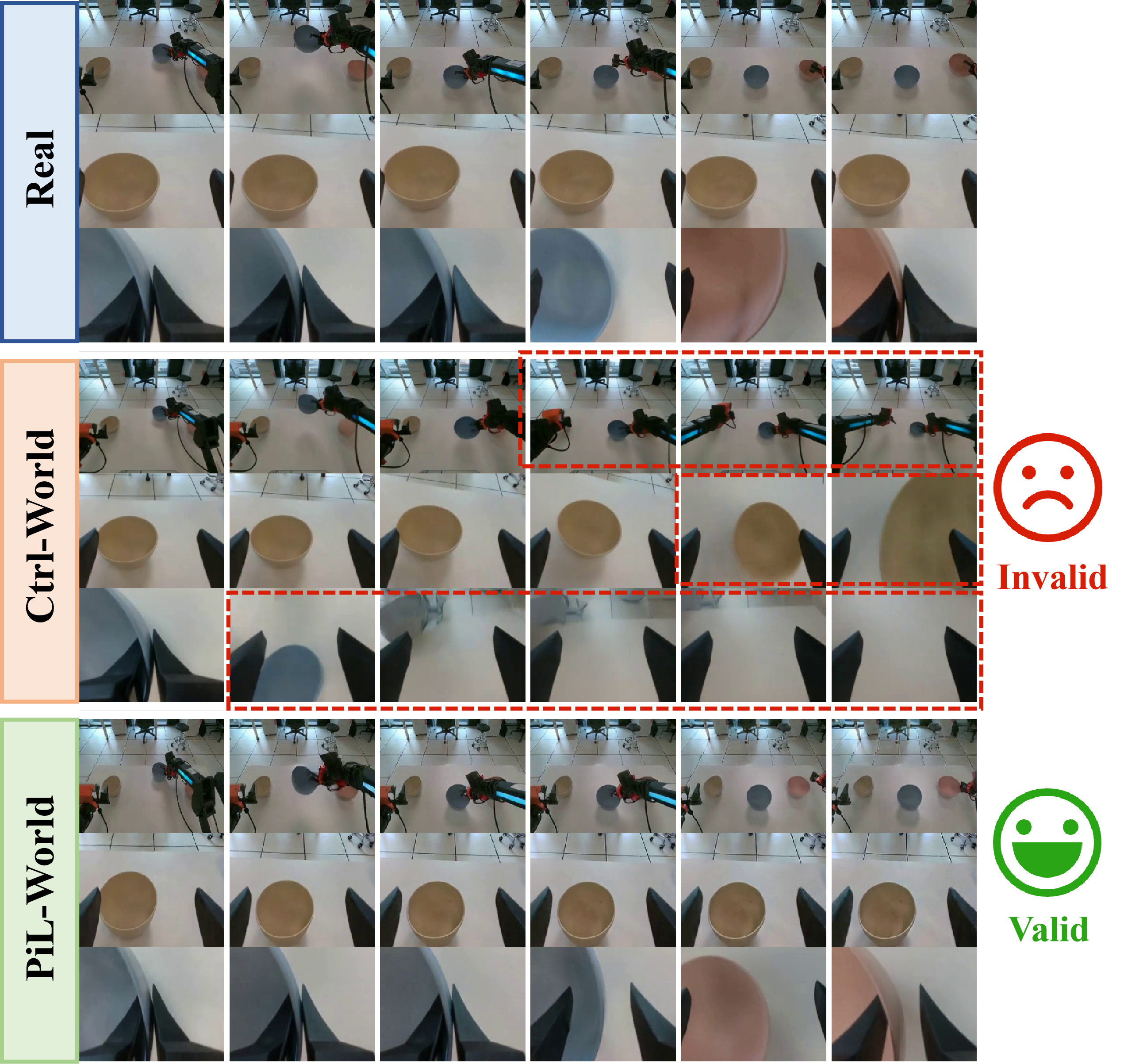}\\[-0.2em]
{\small (b) Rollout example on Stack Bowls}
\end{minipage}
\captionof{figure}{Qualitative rollout examples comparing PiL-World with Ctrl-World on the same rollout segments. Red dashed boxes highlight visible inconsistencies with the reference execution.}
\label{fig:rollout_examples}
\end{center}

\section{Conclusion}
We presented PiL-World, a chunk-wise world model for policy-in-the-loop VLA evaluation. PiL-World generates multi-view future observations from VLA-predicted action chunks and uses the final generated observation as the input to the next policy query, enabling closed-loop imagined rollouts for policy-in-the-loop evaluation. This process composes multiple chunks into a closed-loop imagined rollout. Across three real dual-arm manipulation tasks, PiL-World improves real-imagined success agreement, hallucination-free rollout reliability, and single-step multi-view perceptual similarity over Ctrl-World. These results suggest that chunk-wise world-model rollouts can serve as a useful evaluation signal for VLA policies before real-robot deployment.

\section{Limitations}
Despite these encouraging results, PiL-World has several limitations. First, we evaluate only three dual-arm manipulation tasks, and broader tasks, object configurations, and robot platforms are needed to better assess generalization. Second, contact-rich manipulation remains challenging, particularly in stacking tasks where small contact errors can lead to large deviations in object motion. Third, imagined rollout success and HFR currently rely on human annotation. In addition, the head-view action-to-control projection may be less effective under severe occlusion or when manipulation is primarily observed from wrist cameras.

\bibliography{example}

\appendix
\renewcommand{\thefigure}{\Alph{section}.\arabic{figure}}
\renewcommand{\thetable}{\Alph{section}.\arabic{table}}
\setcounter{figure}{0}
\setcounter{table}{0}

\section{Supplementary Materials}

\subsection{Target-Task Descriptions}
\label{app:task_descriptions}
\begin{figure}[h]
\centering
\includegraphics[width=\linewidth]{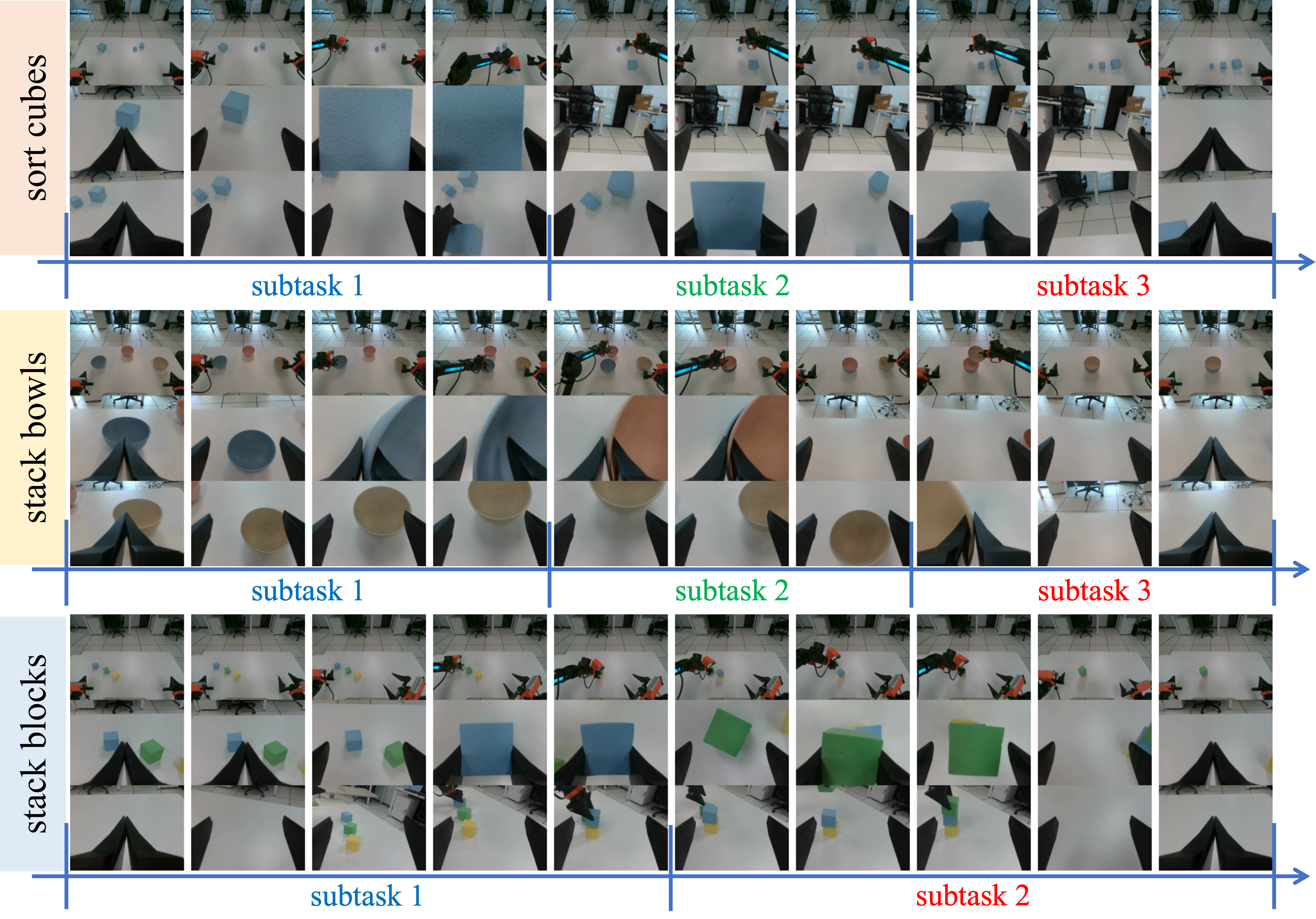}
\caption{Target-task examples and subtask segmentation for Sort Cubes, Stack Bowls, and Stack Blocks.}
\label{fig:task_intro}
\end{figure}

PiL-World is evaluated on three real dual-arm manipulation tasks. Figure~\ref{fig:task_intro} provides visual examples of the target tasks and their subtask segmentation. Closed-loop rollout evaluation is performed at the subtask level: each rollout starts from a subtask state and is paired with the same language instruction used by the VLA policy. Single-step prediction evaluation is performed at the clip level, using extracted trajectory clips with ground-truth action conditions. Table~\ref{tab:target_task_splits} reports the resulting target-task split sizes.

\begin{table}[ht]
\centering
\caption{Target-task split statistics used for fine-tuning and evaluation. Subtask counts correspond to closed-loop rollout evaluation, while clip counts correspond to world-model fine-tuning and single-step prediction evaluation after sliding-window conversion.}
\label{tab:target_task_splits}
\footnotesize
\begin{tabular}{lccccc}
\hline
Task & Train traj. & Train clips & Test traj. & Test subtasks & Test clips \\
\hline
Sort Cubes & 100 & 8707 & 20 & 60 & 1830 \\
Stack Bowls & 100 & 7342 & 18 & 54 & 1301 \\
Stack Blocks & 200 & 15626 & 20 & 40 & 820 \\
\hline
\end{tabular}
\end{table}

\noindent\textbf{Sort Cubes.}
A size-ordering task with three cubes of different sizes. The robot must place the cubes from smallest to largest. We split the episode into three placement subtasks, one for each cube, where the target position is the next empty slot in the ordered layout.
\emph{Instruction:} ``sort the cubes by size.''

\noindent\textbf{Stack Bowls.}
A three-bowl stacking task. Each subtask starts before grasping a bowl and ends after the bowl has been placed at the stacking location. Success requires a stable nested stack rather than only moving the bowl near the target.
\emph{Instruction:} ``pick up the bowls and stack them one by one.''

\noindent\textbf{Stack Blocks.}
A precision stacking task with three blocks. We evaluate two subtasks: placing the second block on the first block, and placing the third block on the resulting two-block stack. The task is sensitive to contact and small pose errors, so the success label requires the tower to remain stable after placement.
\emph{Instruction:} ``pick up the blocks and stack them one by one.''

\subsection{Clip-Level Training Sample Construction}
\label{app:training_clips}

World-model training is performed on clips rather than full episodes. One prediction call covers $K\Delta=15\times3=45$ original time steps. We slide a window over each trajectory and sample $K=15$ future frames at stride $\Delta=3$. Adjacent windows are separated by $9$ stride-sampled frames. Each training sample contains the recent multi-view history, the current observation, the stride-aligned action segment, and the corresponding future frames. 

\subsection{Control-Video Projection Details}
\label{app:control_video}

\begin{figure}[h]
\centering
\includegraphics[width=\linewidth]{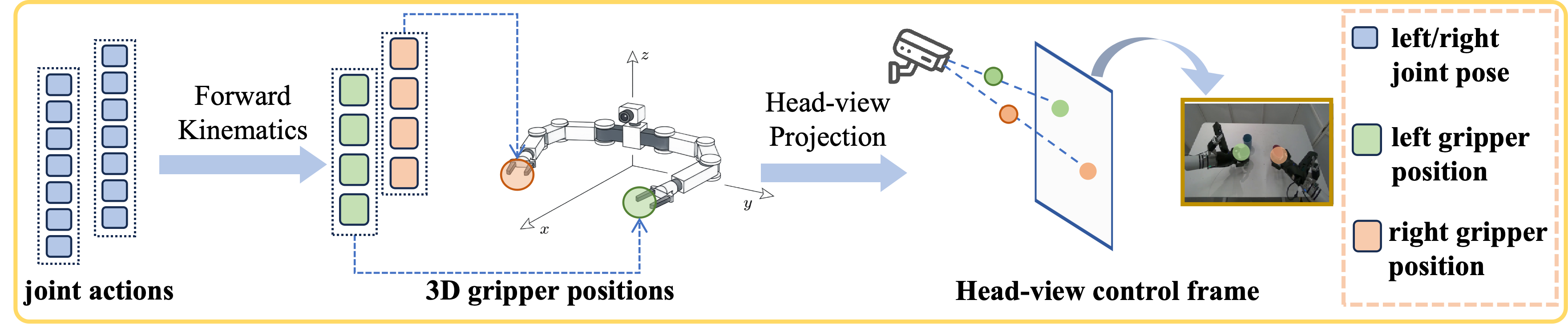}
\caption{Action-to-control projection pipeline. Actions are converted to gripper positions by forward kinematics, projected into the head-view image plane, and encoded as marker-based visual action conditions.}
\label{fig:action_projection}
\end{figure}

The control video is constructed directly from the actions, as illustrated in Figure~\ref{fig:action_projection}. For each action $a_{t+k\Delta}$, we use the absolute joint-position commands to recover the left and right gripper positions by forward kinematics in the robot base frame. The resulting 3D gripper points are projected into the head-view camera using the calibrated intrinsics and extrinsics.

The projected gripper locations are drawn as markers on the control frames. Marker size encodes the gripper open--close state, and short fading traces between adjacent projected positions encode motion continuity. This head-view sequence is used as the visual action condition for generation.

\subsection{Latent Training Objective}
\label{sec:latent_objective}

Let $Z_t^0=\{Z_t^{v,0}\}_{v\in\mathcal{V}}$, $Z_t^h=\{Z_t^{v,h}\}_{v\in\mathcal{V}}$, and $Z_t^f=\{Z_t^{v,f}\}_{v\in\mathcal{V}}$ denote the current-frame, history, and future latents across views. The history memory in the main text is $\mathcal{H}_t=\{Z_t^{v,h}\}_{v\in\mathcal{V}}$. Given the head-view control video $C_t=\Gamma(A_t^{\Delta,K})$ and instruction $g$, the model conditions on $Z_t^0$, $Z_t^h$, $C_t$, and $g$. Only $Z_t^f$ is noised and supervised.

For a sampled noise level $\lambda$ and Gaussian noise $\epsilon$, only the future latents are perturbed as
\[
\tilde{Z}_{t,\lambda}^f = q_\lambda(Z_t^f,\epsilon),
\]
where $q_\lambda$ denotes the video backbone's noising schedule. The current frame, history, control video, and instruction remain clean conditioning inputs, which we group as $\mathcal{C}_t=(Z_t^0,Z_t^h,C_t,g)$. With $\Psi_\theta$ denoting the latent prediction network, the model predicts a backbone-specific target $u_{t,\lambda}$ for the sampled noise level. Depending on the video backbone parameterization, $u_{t,\lambda}$ is the added noise, velocity, or an equivalent denoising target. The training loss is
\[
\mathcal{L}_{\mathrm{gen}}
=
\mathbb{E}_{t,\lambda,\epsilon}
\left[
\left\|
\Psi_{\theta}(\tilde{Z}_{t,\lambda}^f,\lambda,\mathcal{C}_t)
-
u_{t,\lambda}
\right\|_2^2
\right].
\]

\subsection{Implementation Details}
\label{app:baseline}

The VLA policy is frozen during evaluation and outputs action chunks of length $H_\pi=50$. Each action is a $14$-dimensional absolute joint-space command, including two entries for the left and right gripper states. PiL-World uses $224 \times 224$ inputs and outputs, matching the VLA image resolution. During pretraining, we train on RealSource World for $2$ epochs using $64$ H20 GPUs. We then train on the target tasks for $20$ epochs using $8$ H20 GPUs.

PiL-World uses $H_h=5$ history frames. Each round predicts $K=15$ future frames with stride $\Delta=3$, and each subtask uses at most $R=5$ closed-loop rounds. Thus, one rollout step uses stride-aligned action conditions spanning $K\Delta=45$ original time steps from the $50$-step VLA action chunk. This matches the training-clip setup described in Appendix~\ref{app:training_clips}.

\subsection{Ablation on Latent History Memory}
\label{app:history_ablation}

\begin{table}[h]
\centering
\caption{Ablation on latent history memory under the same single-step LPIPS protocol.
Lower LPIPS is better.}
\label{tab:history_ablation}
\footnotesize
\begin{tabular}{lcccc}
\hline
Task & Method & Overall $\downarrow$ & Head $\downarrow$ & Wrist Avg. $\downarrow$ \\
\hline
\multirow{2}{*}{Sort Cubes}
& PiL-World & \textbf{0.0965} & \textbf{0.0597} & \textbf{0.1148} \\
& w/o latent history memory & 0.3146 & 0.3176 & 0.3131 \\
\hline
\multirow{2}{*}{Stack Bowls}
& PiL-World & \textbf{0.1100} & \textbf{0.0597} & \textbf{0.1351} \\
& w/o latent history memory & 0.2759 & 0.3333 & 0.2472 \\
\hline
\multirow{2}{*}{Stack Blocks}
& PiL-World & \textbf{0.1208} & \textbf{0.0617} & \textbf{0.1503} \\
& w/o latent history memory & 0.3403 & 0.3473 & 0.3367 \\
\hline
\end{tabular}
\end{table}

Table~\ref{tab:history_ablation} reports a single-step LPIPS ablation where only the latent history memory is removed. LPIPS is computed between predicted and ground-truth future frames using ground-truth action conditions. Overall LPIPS rises from $0.0965$ to $0.3146$ on Sort Cubes, from $0.1100$ to $0.2759$ on Stack Bowls, and from $0.1208$ to $0.3403$ on Stack Blocks. The increase appears in both the head view and wrist views, which is consistent with the role of recent multi-view history in keeping the generated segment aligned with the preceding rollout context.

\begin{table}[t]
\centering
\caption{LPIPS results on RealSource tasks. Lower values are better.}
\label{tab:realsource_lpips}
\resizebox{\linewidth}{!}{
\begin{tabular}{ccccc}
\hline
\multicolumn{1}{c}{Task} & Top View & Left-wrist View & Right-wrist View & Overall Mean \\
\hline
\emph{Place the slippers} & 0.132 & 0.248 & 0.342 & 0.241 \\
\emph{Make toast} & 0.140 & 0.286 & 0.231 & 0.219 \\
\emph{Hang out the clothes to dry} & 0.187 & 0.232 & 0.217 & 0.212 \\
\emph{Organize the repair tools} & 0.179 & 0.038 & 0.398 & 0.205 \\
\emph{Place the hairdryer} & 0.117 & 0.230 & 0.243 & 0.197 \\
\emph{Clean the convenience store} & 0.124 & 0.230 & 0.236 & 0.197 \\
\emph{Organize the TV cabinet} & 0.143 & 0.253 & 0.193 & 0.196 \\
\emph{Pack the badminton shuttlecock} & 0.149 & 0.326 & 0.071 & 0.182 \\
\emph{Replace the tissues and arrange them} & 0.126 & 0.200 & 0.219 & 0.182 \\
\emph{Take out the trash} & 0.230 & 0.215 & 0.092 & 0.179 \\
\emph{Collect the mail} & 0.171 & 0.028 & 0.315 & 0.171 \\
\emph{Tidy up the kitchen counter} & 0.096 & 0.217 & 0.199 & 0.170 \\
\emph{Organize the glass tube on the rack} & 0.107 & 0.177 & 0.226 & 0.170 \\
\emph{Put the milk in the refrigerator} & 0.136 & 0.057 & 0.315 & 0.169 \\
\emph{Replenish tea bags} & 0.160 & 0.032 & 0.315 & 0.169 \\
\emph{Arrange the cups} & 0.093 & 0.210 & 0.203 & 0.169 \\
\emph{Place the books} & 0.126 & 0.340 & 0.039 & 0.168 \\
\emph{Steam buns} & 0.088 & 0.252 & 0.158 & 0.166 \\
\emph{Cable plugging} & 0.108 & 0.027 & 0.360 & 0.165 \\
\emph{Making steamed potatoes} & 0.075 & 0.212 & 0.207 & 0.165 \\
\emph{Organize the toys} & 0.160 & 0.042 & 0.288 & 0.164 \\
\emph{Tidy up the children's room} & 0.091 & 0.186 & 0.203 & 0.160 \\
\emph{Move industrial parts to different plastic boxes} & 0.068 & 0.029 & 0.376 & 0.157 \\
\emph{Organize the magazines} & 0.075 & 0.308 & 0.083 & 0.155 \\
\emph{Arrange the items on the conference table} & 0.087 & 0.039 & 0.338 & 0.155 \\
\emph{Tidy up the conference room table} & 0.072 & 0.244 & 0.132 & 0.150 \\
\emph{Prepare the birthday cake} & 0.072 & 0.175 & 0.200 & 0.149 \\
\emph{Stack the cups} & 0.072 & 0.161 & 0.211 & 0.148 \\
\emph{Steaming rice in a rice cooker} & 0.066 & 0.197 & 0.180 & 0.148 \\
\emph{Organize the pen holder} & 0.088 & 0.214 & 0.124 & 0.142 \\
\emph{Take down the book} & 0.092 & 0.309 & 0.025 & 0.142 \\
\emph{Tidy up the cooking counter} & 0.104 & 0.258 & 0.028 & 0.130 \\
\emph{Refill the laundry detergent} & 0.047 & 0.289 & 0.045 & 0.127 \\
\emph{Cook rice using an electric rice cooker} & 0.055 & 0.272 & 0.025 & 0.118 \\
\emph{Prepare the bread} & 0.059 & 0.235 & 0.027 & 0.107 \\
\hline
\end{tabular}
}
\end{table}

\begin{figure}[t]
  \centering
  \makebox[\linewidth][c]{\includegraphics[width=\linewidth]{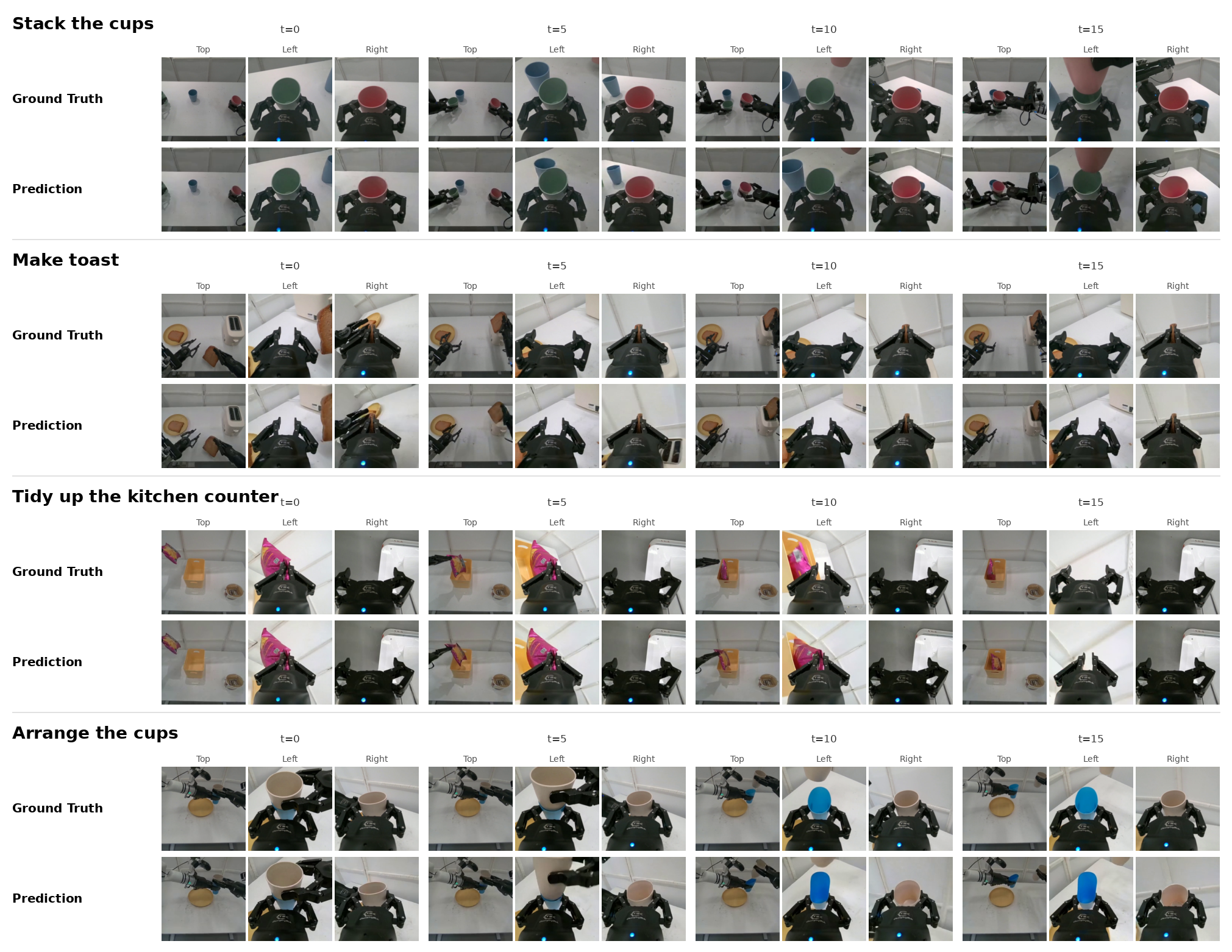}}
  \caption{Qualitative RealSource World single-step prediction examples. Each example compares ground-truth and predicted future observations across synchronized top, left-wrist, and right-wrist views, illustrating the multi-view perceptual similarity of the action-conditioned world model.}
  \label{fig:realsource_qualitative}
\end{figure}

\subsection{RealSource World Results}
\label{app:realsource}

Before pretraining, we reserve $5$ episodes from each RealSource World task for validation. We evaluate single-step action-conditioned prediction on this held-out split. Table~\ref{tab:realsource_lpips} reports LPIPS by task and camera view, and Figure~\ref{fig:realsource_qualitative} shows multi-view prediction examples.

\subsection{Metric Definitions and Judge Protocol}
\label{app:metrics}

\noindent\textbf{Success-rate agreement.}
Let the test set contain $N$ subtask samples, and let each execution be labeled by $M=3$ human judges. For the real-robot execution, let $y_{\mathrm{real}}^{(i,m)}\in\{0,1\}$ denote whether judge $m$ labels the $i$-th subtask as successful. For the world-model prediction, let $y_{\mathrm{imag}}^{(i,m)}\in\{0,1\}$ denote whether judge $m$ labels the predicted rollout for the same subtask as successful. We compute the real-robot success rate and the imagined success rate as
\[
\mathrm{SR}_{\mathrm{real}}
=
\frac{1}{M}
\sum_{m=1}^{M}
\frac{1}{N}
\sum_{i=1}^{N}
y_{\mathrm{real}}^{(i,m)},
\quad
\mathrm{SR}_{\mathrm{imag}}
=
\frac{1}{M}
\sum_{m=1}^{M}
\frac{1}{N}
\sum_{i=1}^{N}
y_{\mathrm{imag}}^{(i,m)}.
\]
The agreement metric is the absolute difference between the two success rates:
\[
|\Delta\mathrm{SR}| =
|\mathrm{SR}_{\mathrm{imag}}-\mathrm{SR}_{\mathrm{real}}|.
\]
A lower $|\Delta\mathrm{SR}|$ means that the world-model rollouts better match the real-robot success rate.

\noindent\textbf{Hallucination-free ratio.}
HFR measures how far an imagined rollout proceeds before the first obvious hallucination. Let $T_i$ be the number of frames in the $i$-th imagined rollout, and let $t_h^{(i,m)}$ be the first hallucination frame annotated by judge $m$. If judge $m$ does not observe an obvious hallucination, we set $t_h^{(i,m)}=T_i$. The metric is
\[
\mathrm{HFR}
=
\frac{1}{M}
\sum_{m=1}^{M}
\frac{1}{N}
\sum_{i=1}^{N}
\frac{t_h^{(i,m)}}{T_i}.
\]
We count a hallucination only when it affects trajectory credibility, such as object disappearance or sudden appearance, cross-view spatial conflict, discontinuous arm or object motion, or non-physical contact and support behavior.

\noindent\textbf{Single-step LPIPS.}
LPIPS is reported only for the single-step protocol. Each model predicts $K=15$ future frames from the same history observations and ground-truth action conditions. Predictions are compared with real future frames aligned at stride $\Delta=3$. We report overall, view-wise, and frame-index-wise LPIPS to separate global perceptual error, camera-specific error, and error growth across the prediction horizon.
\end{document}